\documentclass[letterpaper]{article} 
\usepackage[submission]{aaai2026}  
\usepackage{times}  
\usepackage{helvet}  
\usepackage{courier}  
\usepackage[hyphens]{url}  
\usepackage{graphicx} 
\usepackage{natbib}  
\usepackage{caption} 
\usepackage{algorithm}

\usepackage{newfloat}
\usepackage{listings}
\DeclareCaptionStyle{ruled}{labelfont=normalfont,labelsep=colon,strut=off} 
\floatstyle{ruled}
\newfloat{listing}{tb}{lst}{}
\floatname{listing}{Listing}
\usepackage{amssymb}
\usepackage{amsmath}
\usepackage{subcaption}
\usepackage{algpseudocode} 
\usepackage{multirow}

\title{CADSpotting: Robust Panoptic Symbol Spotting on Large-Scale CAD Drawings}
\author {
    Fuyi Yang\textsuperscript{\rm 1,2\dag},
    Jiazuo Mu\textsuperscript{\rm 1,2\dag},
    Yanshun Zhang\textsuperscript{\rm 2},
    Mingqian Zhang\textsuperscript{\rm 1,2},
    Junxiong Zhang\textsuperscript{\rm 1,2},
    Yongjian Luo\textsuperscript{\rm 1},
    Lan Xu\textsuperscript{\rm 1},
    Jingyi Yu\textsuperscript{\rm 1},
    Yujiao Shi\textsuperscript{\rm 1\ddag},
    Yingliang Zhang\textsuperscript{\rm 2\ddag},
}
\affiliations {
    \textsuperscript{\rm 1}ShanghaiTech University\\
    \textsuperscript{\rm 2}DGene Digital Technology \\
}

\usepackage{bibentry}

\begin{document}

\twocolumn[{
\renewcommand\twocolumn[1][]{#1}
\maketitle
\begin{center}
    \captionsetup{type=figure}
    \includegraphics[width=1\textwidth]{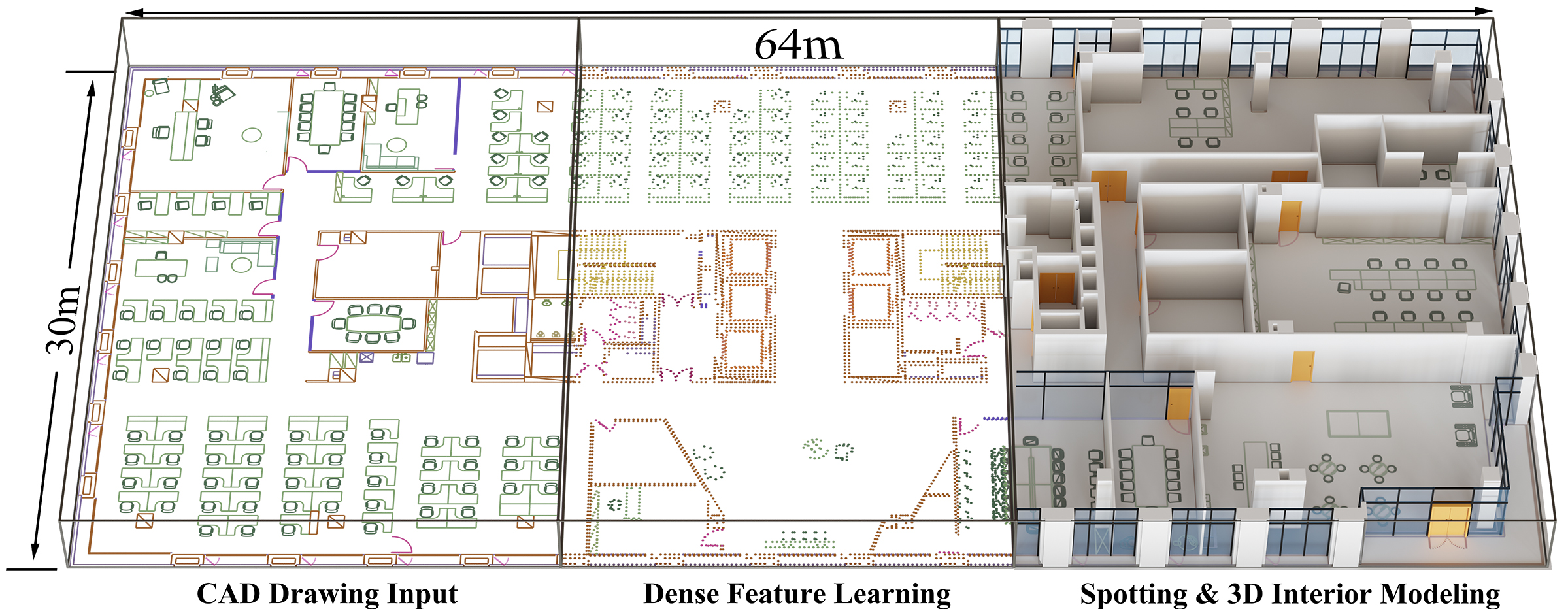}
    \captionof{figure}{CADSpotting accurately identifies and segments symbols in CAD drawings, facilitating tasks like 3D interior modeling. It uses dense point sampling within a unified point cloud model to learn robust primitive features, and integrates Sliding Window Aggregation for efficient panoptic symbol spotting in large-scale drawings. The resulting semantic information enables automated parametric reconstruction of architectural 3D interiors.}
    \label{fig:teasor}
\end{center}
}]
{\let\thefootnote\relax\footnote{$\dag$ Equal contributions. }}\par
{\let\thefootnote\relax\footnote{$\ddag$ Corresponding author. }}\par
\begin{abstract}

We introduce CADSpotting, an effective method for panoptic symbol spotting in large-scale architectural CAD drawings. Existing approaches often struggle with symbol diversity, scale variations, and overlapping elements in CAD designs, and typically rely on additional features (e.g., primitive types or graphical layers) to improve performance. CADSpotting overcomes these challenges by representing primitives through densely sampled points with only coordinate attributes, using a unified 3D point cloud model for robust feature learning. To enable accurate segmentation in large drawings, we further propose a novel Sliding Window Aggregation (SWA) technique that combines weighted voting and Non-Maximum Suppression (NMS). Moreover, we introduce LS-CAD, a new large-scale dataset comprising 45 finely annotated floorplans, each covering approximately 1,000 $m^2$, significantly larger than prior benchmarks. LS-CAD will be publicly released to support future research. Experiments on FloorPlanCAD and LS-CAD demonstrate that CADSpotting significantly outperforms existing methods. We also showcase its practical value by enabling automated parametric 3D interior reconstruction directly from raw CAD inputs.
\end{abstract}

\section{Introduction}
\label{sec:intro}
Architectural Computer-Aided Design (CAD) drawings, especially those used for interior design, serve as detailed digital representations that convey essential information about a building's structure, layout, and intricate details. These drawings include critical elements such as furniture placement, electrical layouts, and spatial organization, ensuring consistency and precision throughout the construction process. 
For instance, the Shanghai Mercedes-Benz Arena utilizes over 3,000 CAD drawings to guide its structural design, optimize spatial arrangements, and streamline the construction workflow. As CAD designs continue to grow in complexity, automating the detection and interpolation of symbols within these floorplans becomes increasingly important 
in various downstream tasks, 
such as code compliance checking and 3D interior modeling~\cite{CAD2Bim}.

Despite the progress made in symbol recognition, the task of panoptic symbol spotting in CAD drawings remains under-explored.
The challenges are multifaceted, arising from the vast diversity of symbol types, the need to differentiate between visually similar symbols, and the presence of overlapping elements. Additionally, variations in scale, orientation, and stylistic representation further complicate accurate identification. 

Traditional query-by-example based methods~\cite{Rusiol2010SymbolSI} struggle with the diverse and complex graphical symbols in real-world CAD drawings. Learning-based approaches~\cite{ren2016faster,redmon2018yolov3} have advanced the field, with early methods using convolutional neural networks (CNNs) or graph-based models~\cite{fan2021floorplancad,Rezvanifar2020SymbolSO,zheng2022gat}, and later works incorporating Transformers and attention mechanisms~\cite{fan2022cadtransformer,zheng2022gat} for improved global reasoning. However, these approaches typically operate on rasterized images derived from vector CAD data, leading to loss of geometric fidelity and recognition errors.
SymPoint~\cite{liu2024spv1} avoids rasterization by treating CAD drawings as point sets and applying point cloud techniques for effective symbol spotting. SymPoint-V2~\cite{liu2024spv2} builds on this by introducing layer-aware encoding and position-guided training. While effective, both methods rely on predefined primitive types, and SymPoint-V2 additionally depends on graphical layer information, limiting their ability to generalize across diverse real CAD data. Moreover, they focus on small to medium-scale drawings and lack mechanisms for preserving spatial consistency in large-scale drawings. As a result, panoptic symbol spotting in expansive CAD drawings remains challenging due to dense clutter, scale variation, and segmentation complexity.

In this paper, we present CADSpotting, a simple yet effective panoptic symbol spotting method tailored for handling CAD drawings at a much larger scale. 
Unlike prior approaches, CADSpotting does not rely on fixed graphic primitive types or additional layer information. Instead, it densely samples points along CAD graphic primitives to construct a compact yet expressive point cloud representation, where each point is represented solely by its coordinate attributes.
A unified 3D point cloud model is employed to learn robust features from these sampled points, followed by a streamlined Transformer decoder that performs panoptic symbol spotting. To handle large-scale, real-world CAD drawings, we introduce a novel Sliding Window Aggregation (SWA) technique that combines weighted voting and Non-Maximum Suppression (NMS) to enable accurate and scalable panoptic segmentation across windowed regions.

On the data front, we further introduce LS-CAD, a new large-scale CAD dataset containing 45 finely annotated floorplans from diverse building types, including campuses and office complexes. Each floorplan spans over 1,000 $m^2$ on average and follows the same fine-grained annotation standards as FloorPlanCAD. LS-CAD is the first dataset of its scale in this domain, and it will be released under a license waiver to support future research. Experimental results on both FloorPlanCAD and LS-CAD demonstrate that CADSpotting outperforms the state-of-the-art methods, demonstrating its robustness and effectiveness in real-world scenarios.
Finally, we demonstrate CADSpotting’s practical value by automating 3D interior reconstruction. Leveraging instance-level outputs, we extract spatial parameters such as wall contours, door/window positions, orientations, and pivot points. These parameters are then used to drive parametric modeling in Blender, enabling efficient and accurate 3D interior generation directly from raw CAD inputs. To our knowledge, this constitutes the first end-to-end pipeline that tightly integrates CAD symbol spotting with parametric 3D reconstruction (Fig.\ref{fig:teasor}).

\section{Related Work}
\label{sec:realted work}
\begin{figure*}
  \centering
    \includegraphics[width=1.0\linewidth]{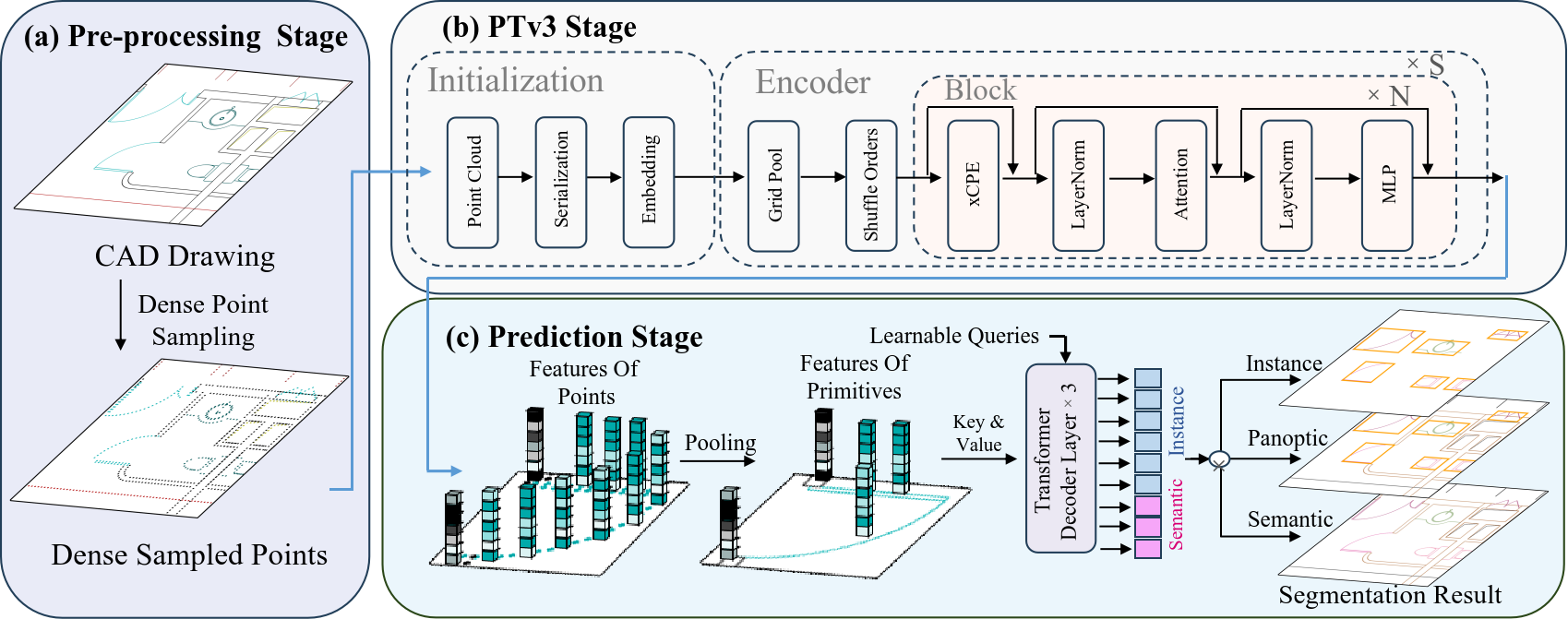}
    \caption{Overview of the CADSpotting pipeline. (a) Given a CAD drawing as input, CADSpotting first densely samples points along CAD graphic primitives to construct a point cloud representation, where each point is defined by its spatial coordinates.(b) Next, PTv3 extracts robust geometric features from the sampled point cloud.(c) Finally, after aggregating primitive-level features via mixed pooling, a streamlined Transformer decoder is employed for efficient panoptic symbol spotting.}
  \label{fig:pipeline}
\end{figure*}

\paragraph{Panoptic Image Segmentation.} Image segmentation is a fundamental task in computer vision, traditionally divided into two main categories: semantic segmentation~\cite{long2015fully,chen2017deeplab,wu2019fastfcn,chen2017rethinking,xie2021segformer} and instance segmentation~\cite{ronneberger2015u,he2017mask,jain2023oneformer,kirillov2023segment}.
While semantic segmentation labels each pixel by class, instance segmentation further distinguishes between object instances of the same class. Recently, SAM~\cite{kirillov2023segment} introduces a large model trained on over one billion masks, leveraging prompt-based interactions for enhanced interactive segmentation. SAM2~\cite{ravi2024sam} extends this approach to video segmentation. 
However, these methods often struggle to represent uncountable background elements (e.g., sky, terrain). To unify countable and uncountable regions, Kirillov et al.~\cite{kirillov2019panoptic} introduce panoptic segmentation, which unifies semantic and instance segmentation, effectively handling both countable and uncountable background components. Early approaches\cite{sun2019hrenet,li2019attentionresnet,chen2020scalingresnet} use CNN backbones, while recent advances like Mask2Former~\cite{cheng2022masked} and OneFormer~\cite{jain2023oneformer} leverage Transformer architectures with mask attention mechanisms for improved accuracy. Large models such as SEEM~\cite{zou2024segment} further refines segmentation using a prompt-driven design. Despite these advances, most methods remain pixel-centric, limiting their effectiveness in processing vector graphics like CAD drawings. As a result, they struggle to capture critical vector-based information, including precise geometry and overlapping relationships.

\paragraph{CAD Symbol Spotting.} Early CAD symbol spotting methods~\cite{nguyen2008symbol,nguyen2009symbol,rezvanifar2019symbol,jiang2021recognizing,yang2023vectorfloorseg} rely on handcrafted feature descriptors, such as shape and structure, coupled with techniques like sliding window searches or graph matching for symbol retrieval. The introduction of deep learning significantly improved performance, with models based on Faster R-CNN~\cite{ren2016faster} and YOLO~\cite{redmon2018yolov3} achieving higher accuracy~\cite{Rezvanifar2020SymbolSO}, though they primarily focus on countable object instances. To address uncountable elements, Fan et al.\cite{fan2021floorplancad} introduce the FloorPlanCAD dataset and a CNN-GCN framework for holistic parsing. Zheng et al.\cite{zheng2022gat} further model CAD drawings as graphs, using adjacency prediction for segmentation. CADTransformer~\cite{fan2022cadtransformer} adopt Vision Transformers to extract features from scalar maps. Inspired by advancements in 3D point cloud segmentation~\cite{zhao2021point,wu2024point,kolodiazhnyi2024oneformer3d,schult2023mask3d}, SymPoint~\cite{liu2024spv1} reformulates CAD drawings as primary point sets and applies point-based learning for effective symbol spotting. SymPoint-V2~\cite{liu2024spv2} extends this with layer-aware encoding and position-guided training. Despite their effectiveness, both methods rely on a fixed set of primitive types, and SymPoint-V2 assumes the availability of reliable layer information. These assumptions often do not hold in real-world CAD data, limiting their ability to capture the full diversity and complexity of CAD symbols.

\section{Method}
\label{sec:method}

An overview of our CADSpotting method is illustrated in Fig.\ref{fig:pipeline}. By incorporating a dense point sampling strategy within a unified point cloud processing model, our method effectively extracts features for CAD primitives, yielding a robust representation that supports various primitive types. We describe each module in detail in the following sections.

\subsection{Primitive Feature Learning}
\label{sec:symbol_feature}
SymPoint~\cite{liu2024spv1} represents CAD primitives as sets of 2D points and constructs handcrafted features for each, encoding attributes such as type (line, arc, circle, ellipse) and length. It then applies point cloud analysis with an Attention with Connection Module to learn primitive representations, followed by a masked-attention Transformer decoder for panoptic symbol spotting. While effective, SymPoint is limited by its reliance on manually defined features and a fixed set of four primitive types, which restricts its ability to handle the full diversity of CAD symbols. Real-world drawings often include complex elements like Bézier curves, making handcrafted features insufficiently robust. To overcome these limitations, we propose a novel primitive representation based on dense point sampling and unified point cloud learning, enabling more generalizable and expressive primitive-level features.

\begin{figure}[t]
  \centering
   \includegraphics[width=.98\linewidth]{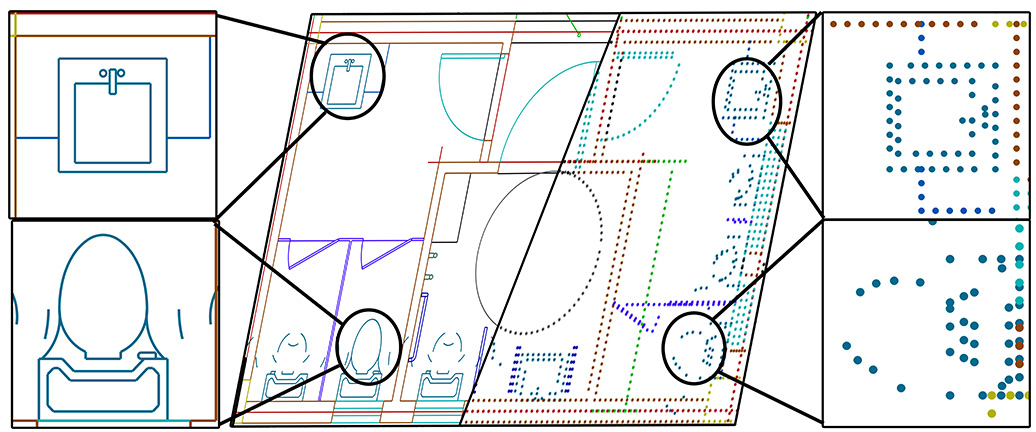}
   \caption{Dense point sampling. Left: A CAD drawing containing various primitives. Right: Dense point sampling converts each primitive symbol to a dense set of points.}
   \label{fig:sampling}
\end{figure}

\paragraph{Dense Point Sampling}

In CAD drawings, graphical primitives are generally classified into line types (e.g., straight segments, curves, Bézier contours) and shape types (e.g., rectangles, circles, ellipses). Inspired by SymPoint~\cite{liu2024spv1}, we adopt a dense point sampling strategy that performs equidistant sampling along each primitive to generate dense point-level data for feature learning. Unlike methods that rely on fixed primitive types and handcrafted features, our sampling approach is more flexible and generalizable, making it suitable for a wide variety of CAD elements. Fig.~\ref{fig:sampling} illustrates the sampling process. Given a CAD drawing with $N$ primitives, we sample points at a fixed interval $d$ to construct a 2D point cloud $P$, where each point $p \in P$ is represented as a 3D vector $(x, y, z) \in \mathbb{R}^3$ with $z=0$. We then use Point Transformer V3~\cite{wu2024point} (PTv3) as the backbone to extract robust features, leveraging its design optimized for unordered point cloud data.

\paragraph{Feature Pooling}

\begin{figure*}[t] 
\centering
\begin{subfigure}[b]{0.32\textwidth} 
\includegraphics[width=\linewidth]{fig/house002_raw_v1.png}
\caption{CAD drawing as raw input}
\label{fig:house002_origin}
\end{subfigure}
\hfill 
\begin{subfigure}[b]{0.32\textwidth}
\includegraphics[width=\linewidth]{fig/house002_pred_2_v1.png}
\caption{Prediction from CADspotting}
\label{fig:house002_pred}
\end{subfigure}
\hfill 
\begin{subfigure}[b]{0.32\textwidth}
\includegraphics[width=\linewidth]{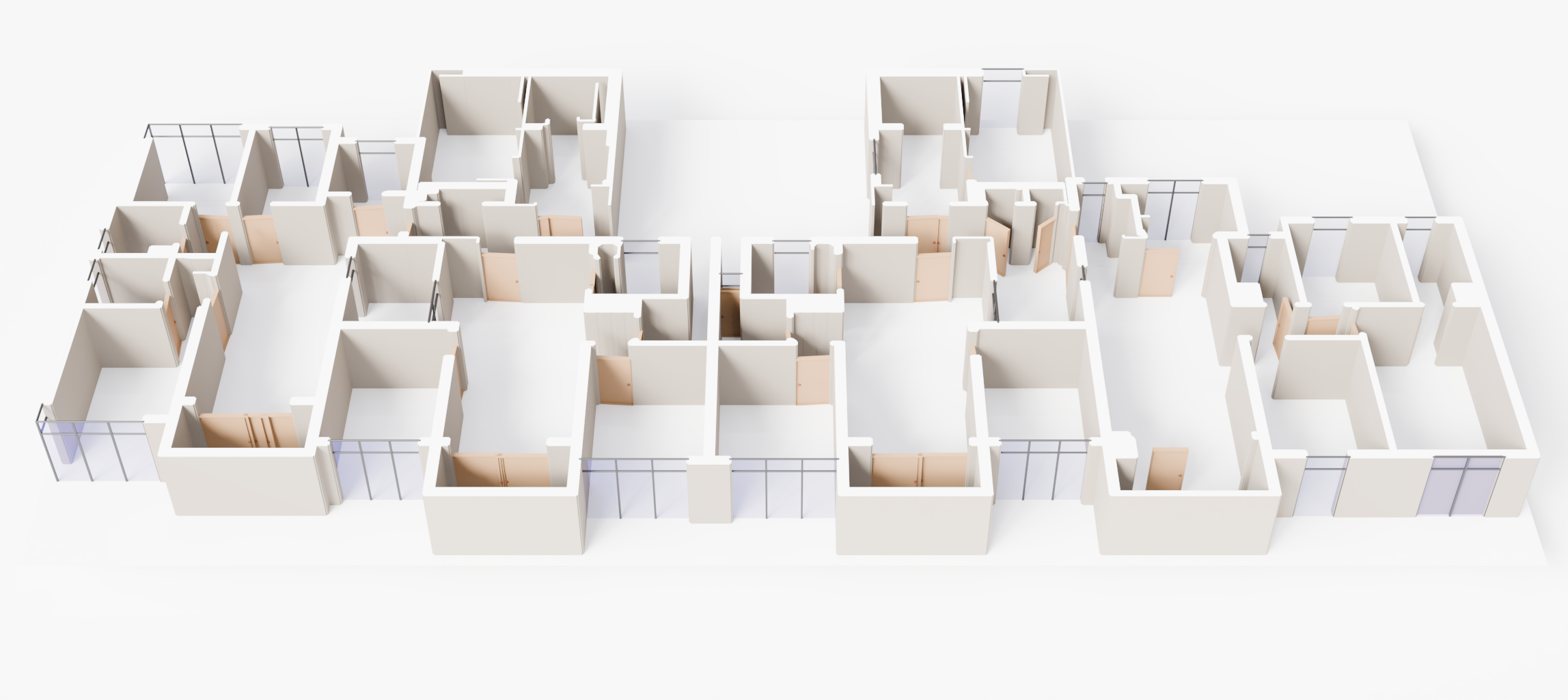}
\caption{Automated reconstruction}
\label{fig:house002_modeling}
\end{subfigure}
\caption{Our CADSpotting takes a CAD drawing as input (a) and outputs predicted semantic and instance segmentation results (b). After that, we design an automated reconstruction pipeline to generate 3D interior models (c).}
\label{fig:modeling}
\end{figure*}

To obtain the final feature representation for CAD primitives, we use Primitive Mixed Pooling (PMP) to convert point-wise features $\mathbf{f} \in \mathbb{R}^{M \times C}$ into primitive-wise features $\mathbf{g}_i \in \mathbb{R}^{N \times C}$.

\begin{align}
    \mathbf{g}_i = \mathcal{PMP}_i(\mathbf{f}) \quad \forall i \in \{1,...,N\} 
\end{align}

where $M$ is the total number of sampled points, and $N$ is the number of primitives. Let $\mathbf{P}_i$ denote the dense point set corresponding to the $i$-th primitive. The $c$-th dimension of the primitive-level feature $\mathbf{g}_i$ is computed as the sum of the maximum and average values of the $c$-th dimension across all points in $\mathbf{P}_i$:

\begin{align}
    \mathbf{g}_i(c) = \mathcal{PMP}_i^c(\mathbf{f}) = \underbrace{\max_{p \in \mathbf{P}_i} \mathbf{f}_p(c)}_{\text{Max pooling}} + \underbrace{\frac{1}{|\mathbf{P}_i|}\sum_{p \in \mathbf{P}_i} \mathbf{f}_p(c)}_{\text{Average pooling}} 
\end{align}

This pooling strategy compresses dense point features into a compact set of primitive-level representations, reducing computational and storage overhead. Meanwhile, it preserves essential geometric information, maintaining global consistency and minimizing information loss. Loss computation is also performed at the primitive level, aligning with the objective of CAD symbol spotting.

\subsection{Panoptic Symbol Spotting}
\label{sec:panoptic_symbol}

We utilize a lightweight Transformer decoder to extract semantic and instance-level information from the pooled primitive features. These features serve as keys and values for three consecutive decoder layers, where self-attention captures dependencies among primitives. Query embeddings are randomly initialized and optimized during training to compute accurate self-attention scores with other features. 

Following OneFormer3D~\cite{kolodiazhnyi2024oneformer3d}, the decoder output is divided into two parts: the first $K_{\text{ins}}$ vectors correspond to instance proposals, and the remaining $K_{\text{sem}}$ vectors represent semantic symbol spotting. The output has shape $N \times (\text{Num}_{\text{ins}} + \text{Num}_{\text{sem}})$, where the first $\text{Num}_{\text{ins}}$ dimensions are used to predict instance masks, and the rest indicate semantic class scores. This unified representation enables panoptic symbol spotting by combining both semantic and instance-level predictions.

Each semantic proposal includes log-likelihoods for each primitive being classified into different semantic labels and can be further processed through thresholding to generate a binary mask. Similarly, an instance proposal also contains log-likelihood values, and these indicate the likelihood of each primitive being classified as a specific instance. 
During training, we apply bipartite matching using the Hungarian algorithm to align predictions with ground truth. At inference, we select the top-$k$ instance proposals with the highest confidence and apply matrix-NMS~\cite{wang2020solov2} to eliminate redundant predictions.

\textbf{Loss Function}
\label{sec:loss_function}
The overall loss function $L$ consists of a classification loss and a primitive mask loss. The classification loss $L_{\text{cls}}$ is defined as a multi-class cross-entropy loss. The mask loss combines binary cross-entropy $L_{\text{bce}}$ and Dice loss~\cite{milletari2016vdiceloss} $L_{\text{dice}}$, jointly measuring the alignment between predicted instance masks and ground truth.

\begin{equation}
\begin{aligned}
L &= \lambda_{\text{cls}} L_{\text{cls}} + \lambda_{\text{bce}}L_{\text{bce}} + \lambda_{\text{dice}} L_{\text{dice}}
\end{aligned}
\end{equation}

\subsection{Sliding Window Aggregation} 
\label{sec:sliding_windows}

To address the challenge of panoptic symbol segmentation at large scales, a key issue is generalizing instance segmentation across drawings of varying sizes. During inference, larger inputs require significantly more instance proposals than those seen during training. However, Mask Transformer-based methods, including ours, are constrained by a fixed number of instance queries. Fortunately, CAD floorplans offer an advantage: instances of the same category typically maintain consistent sizes across different scenes under uniform scaling. This allows us to assume that instance sizes at inference do not exceed those in training data.

Based on this assumption, we adopt a fixed-size sliding window strategy, using windows aligned with training dimensions to process large input drawings. For each window, we collect all primitives it covers or intersects and apply the decoder to generate local semantic and instance proposals. Aggregated results across all windows are then merged using distinct strategies for semantic and instance outputs.

For semantic proposals, we apply a weighted voting scheme to assign each primitive a class label. When a primitive appears in multiple windows, we prioritize proposals from windows that observe a larger portion of it. The voting weight is proportional to the number of observed dense points relative to the total points of the primitive.

For instance proposals, we use Sparse Non-Maximum Suppression (Sparse-NMS) to merge results from overlapping windows efficiently (see supplementary material). By using a sufficiently small step size, each instance is fully captured in at least one window. Incomplete detections from other windows are eliminated during the NMS process.

\subsection{Automated 3D Interior Reconstruction}
\label{sec:auto_reconstruction}

CADSpotting produces accurate panoptic segmentation results that serve as the foundation for our parametric 3D interior modeling pipeline. Using instance-level segmentation and primitive positions, we extract spatial parameters for key architectural elements, including walls, doors, and windows. For doors, we compute positions, orientations, and pivot points by analyzing geometric relationships between arcs and lines within each segmented instance. Window positions are directly derived from their corresponding instances. Wall modeling involves a more specialized process: we merge endpoints of adjacent lines within wall instances to form closed polygons, which are rasterized into binary masks. Final wall contours are then extracted using the method from~\cite{SUZUKI198532}. These spatial parameters are fed into a parametric reconstruction pipeline implemented in Blender. Predefined 3D assets for walls, doors, and windows are used as reusable templates, enabling efficient and automated generation of structurally accurate 3D interiors. This approach converts raw CAD inputs into detailed 3D models within minutes. An example result is shown in Fig.~\ref{fig:modeling}, with further details provided in the supplementary material.

\begin{table}[]
\centering
\begin{tabular}{c|ccc}
\hline

Building & Area($m^2$) & \#Primitive & \#Instance \\ \hline
Office 1  &   1,799   & 28,193      & 1,185  \\ 
Campus 1  & 1,949           & 12,571        & 603        \\ 
Hotel 1 &   1,193   &  26,709  &   335        \\ 

\hline
\end{tabular}
\caption{Summary statistics for three LS-CAD samples.}
\label{tab:our_dataset}
\end{table}

\section{Experiments}
\subsection{The Proposed LS-CAD Dataset}

We introduce LS-CAD, a new large-scale CAD dataset comprising 45 floorplans from expansive buildings, such as campuses and office complexes. Each drawing covers at least 1,000 square meters, with the number of primitives ranging from approximately 2,900 to over 10,000. Table.~\ref{tab:our_dataset} presents three representative examples, detailing area, primitive count, and instance count to highlight the large-scale nature of LS-CAD. The dataset features a rich variety of complex symbols, including doors, windows, walls, elevators, and parking spaces. All floorplans are annotated with fine-grained labels consistent with the FloorPlanCAD~\cite{fan2021floorplancad} standard. The LS-CAD dataset will be publicly released to support future research in panoptic symbol spotting for large-scale CAD drawings.

\subsection{Experimental Settings}
\paragraph{Implementation Details} 
The FloorPlanCAD dataset comprises approximately 15,000 drawings, annotated across 35 symbol categories, including 30 things and 5 stuff classes. To ensure fair comparison, we utilize the same training subset from the FloorPlanCAD dataset as used by SymPoint. 
We train our model on a machine with 8 NVIDIA A100 GPUs, using a batch size of 2 per GPU for 512 epochs. For dense point sampling, we set a fixed distance $d = 0.14$.
We use the AdamW~\cite{loshchilov2017decoupled} optimizer with a learning rate of $10^{-4}$ and a weight decay of 0.05. Data augmentation techniques include random horizontal flipping with a probability of $0.5$, global rotation and scaling transformations, translation along the x and y axes. We set loss weight as
$ \lambda_{\text{cls}} \colon \lambda_{\text{bce}}  \colon \lambda_{\text{dice}} = 0.5  \colon 1  \colon 1 $
. Additionally, we set \( K_{\text{sem}} = 36 \) as the number of semantic labels, and both the top-\( k \) value and \( K_{\text{ins}}\) are set to 220. We use $ \Delta = 70 $ as the step size of SWA on LS-CAD dataset.

\paragraph{Metrics}
Following the approach of~\cite{fan2022cadtransformer}, we assess the performance of our model using multiple metrics. For semantic symbol spotting, we use F1 score and weighted F1 score (wF1). For instance symbol spotting, we employ AP50, AP75, and mean Average Precision (mAP). For panoptic symbol spotting, we utilize Panoptic Quality (PQ), Segmentation Quality (SQ), and Recognition Quality (RQ). PQ is a metric combining SQ and RQ to evaluate model performance in semantic and instance segmentation tasks. Further details on metric formulation are available in~\cite{fan2022cadtransformer}.

\begin{table}[]
\centering
\setlength{\tabcolsep}{10pt}
\begin{tabular}{c|c|c}
\hline
Method                    & F1            & wF1                     \\ \hline
PanCADNet~\cite{fan2021floorplancad}                       & 80.6          & 79.8                     \\
CADTransformer~\cite{fan2022cadtransformer}              & 82.2          & 90.1                    \\
GAT-CADNet~\cite{zheng2022gat}                   & 85.0          & 82.3             \\
SymPoint~\cite{liu2024spv1}                              & 86.8          & 85.5               \\ \hline
\textbf{CADSpotting (ours)}    & \textbf{92.8} & \textbf{93.2} \\ \hline
\end{tabular}
\caption{Quantitative comparison of semantic symbol spotting on FloorPlanCAD dataset.}
\label{tab:sem_comparison}
\end{table}

\subsection{Quantitative Evaluation}

\begin{table}[]
\centering
\setlength{\tabcolsep}{9pt}
\begin{tabular}{c|c|c|ccc}
\hline
Method                    & AP50          & AP75          & mAP           \\ \hline
DINO~\cite{zhang2022dino}                       & 64.0          & 54.9          & 47.5          \\
SymPoint~\cite{liu2024spv1}                      & 66.3          & 55.7          & 52.8          \\
\hline
\textbf{CADSpotting (ours)} & \textbf{70.6} & \textbf{66.6} & \textbf{66.3} \\ \hline
\end{tabular}
\caption{Quantitative comparison of instance symbol spotting on FloorPlanCAD dataset.}
\label{tab:ins_comparison}
\end{table}

\begin{table*}[]
\centering
\begin{tabular}{c|ccc|ccc|ccc}
\hline
\multirow{2}{*}{Method} & \multicolumn{3}{c|}{Total}  & \multicolumn{3}{c|}{Thing} & \multicolumn{3}{c}{Stuff}                     \\ \cline{2-10} 
                               & PQ   & SQ   & RQ            & PQ      & SQ      & RQ     & PQ            & SQ            & RQ            \\ \hline
PanCADNet~\cite{fan2021floorplancad}             & 59.5 & 82.6 & 66.9          & 65.6    & 86.1    & 76.1   & 58.7          & 81.3          & 72.2          \\
CADTransformer~\cite{fan2022cadtransformer}      & 68.9 & 88.3 & 73.3          & 78.5    & 94.0    & 83.5   & 58.6          & 81.9          & 71.5          \\
GAT-CADNet~\cite{zheng2022gat}                   & 73.7 & 91.4 & 80.7          & -       & -       & -      & -             & -             & -             \\
SymPoint~\cite{liu2024spv1}                      & 83.3 & 91.4 & 91.1          & 84.1    & \textbf{94.7}    & 88.8   & 48.2          & 69.5          & 69.4          \\
 \hline
\textbf{CADSpotting (Ours)} & \textbf{87.4} & \textbf{93.5} & \textbf{93.4} & \textbf{88.3} & 94.2 & \textbf{93.7} & \textbf{71.5} & \textbf{82.9} & \textbf{86.3} \\ \hline
\end{tabular}
\caption{Quantitative comparison of panoptic symbol spotting on FloorPlanCAD dataset. 
Our method achieves the best performance among the comparison algorithms.
}
\label{tab:pq_comparison}
\end{table*}

To demonstrate the effectiveness of our approach, we compare it against SOTA methods on the FloorPlanCAD~\cite{fan2021floorplancad} and LS-CAD datasets. A detailed comparison of semantic, instance, and panoptic symbol spotting tasks is provided in the following paragraphs.

\textbf{Semantic Symbol Spotting}
As shown in Table.\ref{tab:sem_comparison}, we compare our method with PanCADNet~\cite{fan2021floorplancad}, CADTransformer~\cite{fan2022cadtransformer}, GAT-CADNet~\cite{zheng2022gat}, and SymPoint~\cite{liu2024spv1}. Our CADSpotting achieves the highest performance in both the F1 and wF1 metrics.

\textbf{Instance Symbol Spotting}
We also evaluate our method against DINO~\cite{zhang2022dino} and SymPoint for instance symbol spotting, using the bounding box maximization of predicted mask to compute the AP metric, consistent with SymPoint. As shown in Table.\ref{tab:ins_comparison}, our method achieves superior results
, with improvements of 4.3 p.p. AP50, 10.9 p.p. AP75 and 13.5 p.p. mAP over SymPoint.

\begin{figure}[t]
  \centering
   \includegraphics[width=1\linewidth]{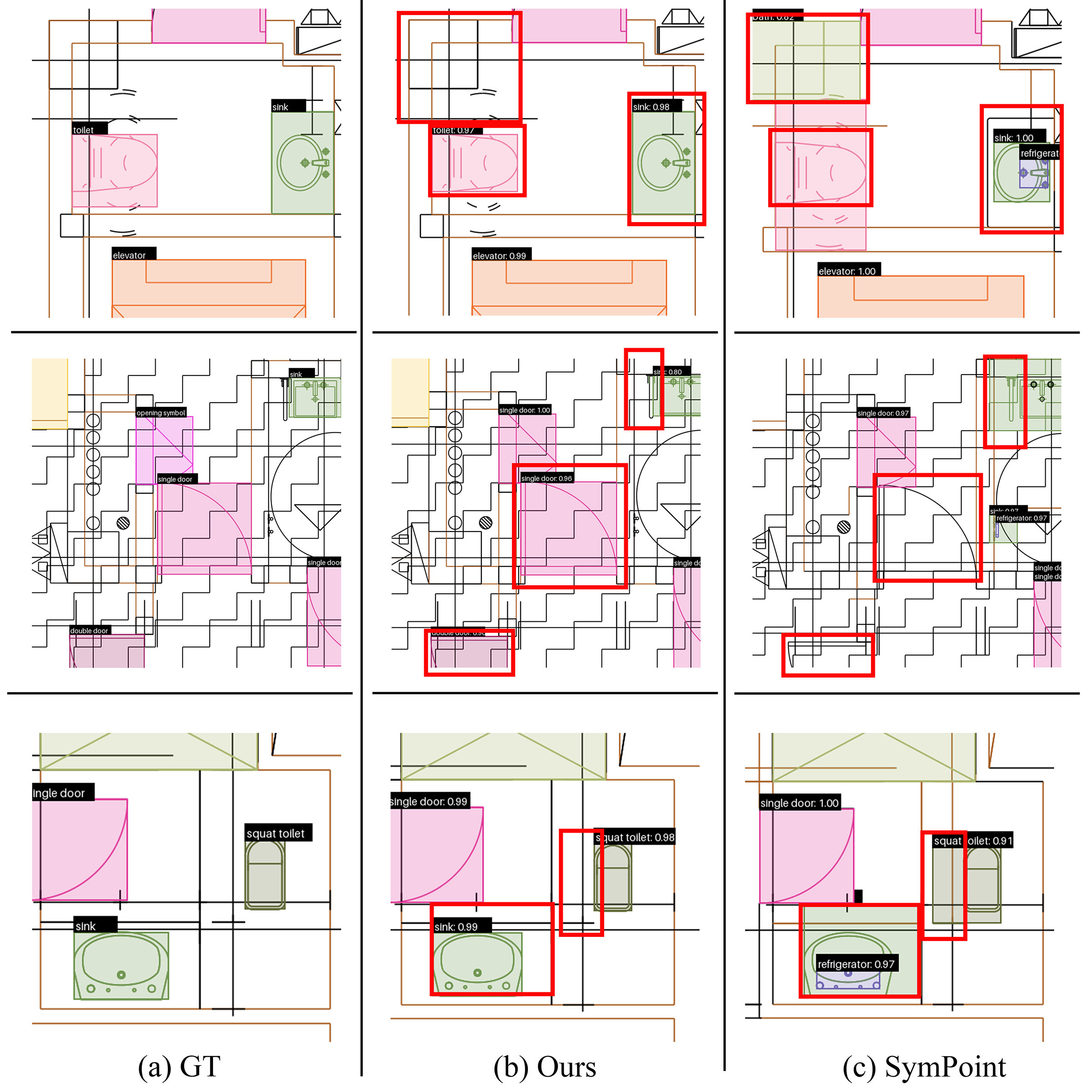}

   \caption{Qualitative comparison of panoptic symbol spotting. 
   }
   
   \label{fig:ins_comparison}
\end{figure}

\textbf{Panoptic Symbol Spotting}
We compare our method with the same methods used in the semantic symbol spotting comparison. The results, presented in Table.\ref{tab:pq_comparison}, show that our approach surpasses SymPoint and other prior methods across all metrics, including Total/Thing/Stuff PQ, SQ and RQ. 
Additionally, our approach offers superior generalization across diverse CAD primitives, thanks to its dense point sampling based feature representation.

\begin{table}[h]
\centering
\footnotesize 
\begin{tabular}{c|ccc|c|c}
\hline
\multirow{2}{*}{Method} & \multicolumn{3}{c|}{Panoptic}  & \multicolumn{1}{c|}{Semantic} & \multicolumn{1}{c}{Instance}        \\ \cline{2-6} 
 & PQ   & SQ   & RQ          & F1                             & mAP        \\ \hline
SPv2                   & \textbf{90.1}	 &\textbf{96.3}	 &\textbf{93.6}           & 89.5                & 60.1                      \\
Ours                 &  {87.4} &  93.5  & 93.4     &\textbf{92.8} &      \textbf{66.3}     \\
 \hline
\end{tabular}
\caption{Quantitative comparison with SPv2 on the FloorPlanCAD dataset.}
\label{tab:sym_comparison}
\end{table}

\begin{table}[h]
\centering
\footnotesize 
\begin{tabular}{c|ccc|c|c}
\hline
 \multirow{2}{*}{Method} & \multicolumn{3}{c|}{Panoptic}  & \multicolumn{1}{c|}{Semantic} & \multicolumn{1}{c}{Instance}        \\ \cline{2-6} 
 & PQ   & SQ   & RQ          & F1                             & mAP                \\ \hline
SPv2    + BP           &26.7   &82.0 &32.6    &55.2 &15.1               \\
Ours   + BP            &58.8 & \textbf{83.7} & 70.2    &  88.7  &       48.3       \\
Ours   + SWA           &\textbf{75.5} & 80.9 & \textbf{93.3}     &\textbf{93.5} &      \textbf{57.5}             \\
 \hline
\end{tabular}
\caption{Quantitative comparison of semantic and panoptic symbol spotting on LS-CAD dataset.}
\label{tab:sliding}
\end{table}

We then compare our method with the concurrent work SymPoint-V2 (SPv2)~\cite{liu2024spv2}, which incorporates primitive types into feature vectors. Our first comparison is conducted on the FloorPlanCAD dataset. As shown in Table.\ref{tab:sym_comparison}, although SPv2 achieves up to 2.7 percentage points higher PQ in panoptic segmentation, our method delivers a 3.3-point gain in F1 score and a 6.2-point improvement in mAP. While SPv2 performs well on FloorPlanCAD due to its reliance on four fixed primitive types, this limits its generalization to more diverse CAD symbol sets. In contrast, our method relies solely on coordinate attributes, enabling more flexible and robust feature learning. Furthermore, SPv2 depends on CAD layer information, which is often inconsistent or unavailable in real-world scenarios. Table.~\ref{tab:sliding} further supports this analysis. On the LS-CAD dataset, where layer annotations are unavailable, our method significantly outperforms SPv2 across all metrics, demonstrating stronger robustness in practical applications.

We also evaluate the effectiveness of Sliding Window Aggregation (SWA) versus Block Partitioning (BP) in CADSpotting on the LS-CAD dataset. In the BP setting, we divide large-scale CAD drawings into non-overlapping blocks matching the dimensions used in FloorPlanCAD. To ensure unique primitive representation, only those with starting points inside a block are retained. As shown in Table~\ref{tab:sliding}, SWA substantially outperforms BP in panoptic quality (PQ), achieving a 16.7-point improvement, with only a modest 2.8-point reduction in segmentation quality (SQ). These results demonstrate that CADSpotting, when combined with SWA, achieves accurate and scalable symbol detection in large-scale CAD drawings.

\begin{figure*}
  \centering
  \setlength{\abovecaptionskip}{0pt}
\setlength{\belowcaptionskip}{0pt}
    \includegraphics[width=1.0\linewidth]{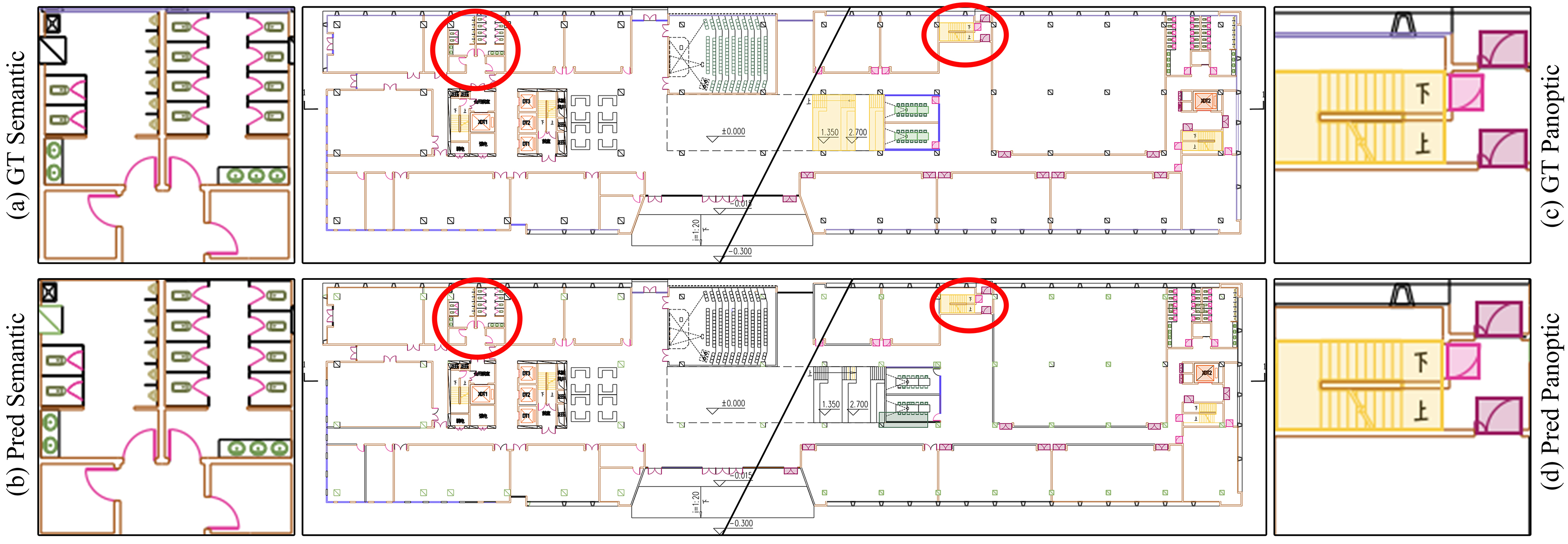}
    \caption{
    Semantic and panoptic symbol spotting results of CADSpotting with the SWA technique on a large-scale office building CAD drawing from our LS-CAD dataset. The figure shows full CAD drawings with GT and predicted results in the center, with close-up views(as indicated by the red ellipses) of GT in (a, c) and predicted results in (b, d).
    }
  \label{fig:LargeScale}
\end{figure*}

\subsection{Qualitative Comparison}
We perform qualitative comparisons with SymPoint on the FloorPlanCAD dataset. 
Fig.\ref{fig:ins_comparison} shows that our method achieves accurate and robust panoptic symbol spotting, even in challenging scenarios involving walls, complex or overlapping symbols, and uncommon furniture symbols. 
This demonstrates the effectiveness of our proposed dense point sampling based primitive features. Additionally, Fig.\ref{fig:LargeScale} presents our semantic and panoptic symbol spotting results on a large-scale CAD drawing of an office building with 28,193 primitives and 1,185 instances from our LS-CAD dataset. These results demonstrate that our CADSpotting enhanced by SWA technique, accurately identifies semantic and instance information in large-scale CAD drawings.

To evaluate model generalizability and demonstrate the utility of LS-CAD dataset, we also train models on a combined dataset consisting of FloorPlanCAD and LS-CAD. Comprehensive experimental results and analyses are provided in the supplementary material.

\begin{table}[]
\centering
\begin{tabular}{c|c|ccc}
\hline
Method                    & Backbone   &PQ          & SQ          & RQ           \\ \hline
Handcrafted Features   & PTv1   & 78.4 & 88.5 & 88.6        \\ \hline
Handcrafted Features   & PTv3   &80.2 & 90.2 & 89.0        \\ \hline
\textbf{Ours} & PTv3   &\textbf{87.4}   & \textbf{93.5}   &
 \textbf{93.4}\\ \hline
\end{tabular}
\caption{Ablation study on feature learning methods on FloorPlanCAD dataset. 
All methods utilize the same decoder architecture to learn intermediate features. 
}
\label{tab:feature ablation}
\end{table}

\subsection{Ablation Studies}
We further present ablation studies on feature learning. Additional ablation studies, which explore various pooling strategies and the use of color information, and evaluate performance across all classes, are provided in the supplementary material.

\textbf{Feature Learning}
We compare our dense point sampling based primitive feature learning method with the approach in SymPoint, which uses handcrafted features to represent each individual primitive. SymPoint employs Point Transformer(PTv1)~\cite{zhao2021point} as its backbone network and incorporates hierarchical multi-resolution primitive features to leverage intermediate features, enhancing the decoder’s performance. To evaluate the impact of different point cloud analysis methods, we first replace SymPoint’s backbone with PTv3~\cite{wu2024point}, maintaining the same decoder architecture as ours to ensure a fair comparison of feature learning techniques in panoptic symbol spotting. As shown in Table.\ref{tab:feature ablation}, with the same decoder configuration, simply switching to PTv3 results in a 1.8 p.p. improvement in PQ. When we apply our proposed primitive feature learning method, performance further increases 7.2 p.p. in PQ, demonstrating the significant performance gains achieved by our dense point sampling based feature learning approach.

\section{Conclusion}
\label{sec:conclusion}

We have introduced CADSpotting, a novel method for panoptic symbol spotting in large-scale CAD drawings. Our approach uses dense point sampling within a unified point cloud framework to represent CAD primitives, combined with a Sliding Window Aggregation technique that integrates weighted voting and Non-Maximum Suppression for precise segmentation. We also present LS-CAD, a dataset of 45 annotated floorplans from diverse building types. Experiments on LS-CAD and FloorPlanCAD datasets have confirmed the effectiveness and scalability of our method, with successful applications in automated 3D reconstruction demonstrating its practical value.

Our approach currently faces limitations due to the dataset's primary emphasis on residential and commercial buildings, which restricts its applicability to industrial and cultural contexts. Additionally,it may struggle to capture significant stylistic variations among CAD symbols. Addressing these challenges through the integration of in-context learning may help enhance annotation efficiency and extend applicability. This is left as our future work.

\bibliography{aaai2026}

\clearpage
\section{Supplementary Material}

\begin{table*}[t]
\setlength{\abovecaptionskip}{0pt}
\setlength{\belowcaptionskip}{0pt}
\centering
\begin{tabular}{c|c|ccc|ccc}
\hline
\multirow{2}{*}{Method} & \multirow{2}{*}{Training Dataset}  & \multicolumn{3}{c|}{Test on FloorPlanCAD} & \multicolumn{3}{c}{Test on Cross-dataset}                     \\ \cline{3-8} 
                      &                 & PQ   & SQ   & RQ             & PQ      & SQ      & RQ                \\ \hline
SymPoint~\cite{liu2024spv1}   & FloorPlanCAD        & 83.3 & 91.4 & 91.1          & 33.2 & 83.0 & 39.9       \\
SymPoint                      & Cross-dataset    & 79.0 & 89.2 & 88.6          & 57.0 & 84.0 & 67.8       \\
CADSpotting                   &  FloorPlanCAD       & {\textbf{87.4}} & {\textbf{93.5}} & {\textbf{93.4}}          & 52.9 & 90.4 & 58.5          \\
CADSpotting                   &  Cross-dataset       & {{84.0}} & {{91.3}} & {{91.9}}       & {58.8} & 83.7 & 70.2         \\
CADSpotting(SWA)              &  FloorPlanCAD        &      &   /   &               & 60.1 &   \textbf{90.9} & 66.1          \\
CADSpotting(SWA)              &  Cross-dataset        &      &  /    &           & \textbf{75.5}  & 80.9 & \textbf{93.3}            \\
 \hline
\end{tabular}
\caption{Cross-dataset joint training performance comparison between SymPoint and CADSpotting. 
\\
}

\label{tab:A+B}
\end{table*}

\section{More Details on the LS-CAD Dataset}
\label{sec:lscaddataset}

In Fig.\ref{fig:LS_raw_gt}, we present visualizations of several samples from the LS-CAD dataset to highlight the diversity of the proposed dataset. The top panel illustrates Office 3, 
which spans a floor area of 2,994$m^2$, including 16,803 primitives and 559 instances. The bottom panel depicts Hotel 1, which spans a floor area of 1,193$m^2$, consisting of 26,709 primitives and 335 instances.


To demonstrate the generalization capability of our model and highlight the value of the LS-CAD dataset, we implement a cross-dataset joint training paradigm. We randomly partition complete large-scale CAD drawings into three subsets: 80\% for training, 10\% for validation, and 10\% for testing. The division of the LS-CAD dataset strictly adheres to the standardized data splitting protocol established in FloorPlanCAD. The training and validation sets from both datasets are subsequently integrated to construct a cross-dataset collaborative training environment. As shown in Table.\ref{tab:A+B}, the cross-dataset joint training strategy significantly improves model performanc on the LS-CAD test set, although with slight performance degradation on the original FloorPlanCAD test set. This demonstrates the necessity of the proposed LS-CAD dataset for provding more diverse CAD representations. 
Notably, CADSpotting+SWA method reaches the best performance of 75.5 on the PQ metric.
After cross-dataset joint training, the PQ score for CADSpotting decreases by only 3.4 p.p. on FloorPlanCAD, whereas it drops by 4.3 p.p. on SymPoint. This indicates that our model exhibits stronger generalization capabilities.


\section{Quantitative Evaluation}
\label{sec:quantity}
\paragraph{Ablation Study on Feature Learning.}  
We present a comprehensive evaluation of panoptic quality (PQ), segmentation quality (SQ), and recognition quality (RQ) for each class, comparing different feature learning methods. Our final approach achieves superior performance across the majority of classes, particularly excelling in commonly used categories such as door classes, window classes, and wall classes. The detailed results are shown in Table.\ref{tab:class_PQ}.

\paragraph{Ablation Study on Pooling.}
We further evaluate the necessity and impact of different pooling methods on our approach, using a baseline configuration without a pooling layer. As shown in Table.\ref{tab:pooling ablation}, max pooling improves performance by 5.0 p.p., average pooling by 5.1 p.p., and mixed pooling by 5.5. p.p. in PQ. Based on these results, we select mixed pooling for our CADSpotting approach.

\paragraph{Ablation Study on Prior Color Information.}
To evaluate the necessity of color prior, we incorporate RGB channels as an optional input. As shown in Table \ref{tab:color ablation}, adding color improves PQ by 1.5 p.p. and SQ by 2.1 p.p., while slightly decreasing RQ by 0.4 p.p. More importantly, the geometry-based representation (\textit{w/o Color}) achieves strong baseline performance, demonstrating that our model maintains high recognition quality even without color information.

\section{Qualitative Evaluation}
\label{sec:quality}
We provide comparisons of prediction results on different datasets. For FloorPlanCAD, we compare the ground truth and prediction using their primitive colors to intuitively evaluate prediction accuracy, as shown in Fig.\ref{fig:qualitative_comparison_1}, Fig.\ref{fig:qualitative_comparison_2} and Fig.\ref{fig:sem_comparison}. For LS-CAD, we present the results of semantic and panoptic predictions, highlighting the performance of the method in practical scenarios, as shown in Fig.\ref{fig:qualitative_comparison_3} and Fig.\ref{fig:qualitative_comparison_4}.

\section{Sliding Window Implementation Details}
\label{sec:slidingwindow}
The Algorithm.\ref{alg:seg_fusion} details the implementation process of our SWA. In the matrix-NMS~\cite{wang2020solov2} process for sliding window aggregation, computing the Intersection over Union (IoU) between all pairs of instance proposal masks is essential for constructing the IoU matrix. However, the number of instance proposals grows substantially with the size of the input drawing. If a dense matrix representation is used, both the time and memory complexities of this operation become prohibitively high.

\begin{figure}[t]
  \centering
  \setlength{\abovecaptionskip}{0pt}
  \setlength{\belowcaptionskip}{0pt}
   \includegraphics[width=1\linewidth]{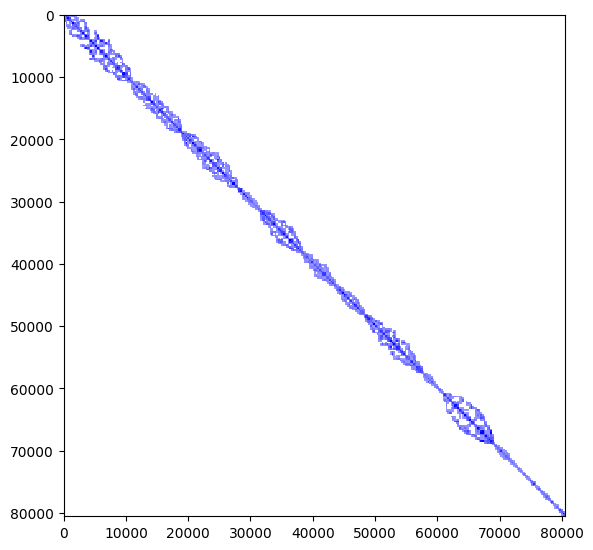}

   \caption{Visualization of the sparsity structure of the IoU matrix during sliding window aggregation for a sample drawing from the LS-CAD dataset. The Reverse Cuthill-McKee (RCM) algorithm was applied to reorder the matrix rows and columns, reducing its bandwidth and improving the visualization.}
   \label{fig:nms_spy}
\end{figure}

To mitigate this issue, we observe that only a small fraction of proposal pairs exhibit nonzero overlap, as most proposals belong to different inference windows. Fig.~\ref{fig:nms_spy} visualizes the sparsity structure of the IoU matrix during sliding window aggregation. By leveraging this sparsity, we reimplemented the IoU computation using sparse matrix routines, significantly reducing computational cost and memory usage without sacrificing accuracy.

\section{Automated 3D Interior Reconstruction}
Based on the spotting results generated by CADSpotting, we present more 3D reconstruction renderings in Fig.\ref{fig:modeling results}. By systematically optimizing the parameters of predefined instance components using instance segmentation data and primitive spatial coordinates, we achieve accurate 3D model reconstruction through component-level geometric adaptation

For wall polygon extraction, we employ an innovative SVG-PNG conversion strategy to differentiate floor areas from walls: 
1) Merge adjacent endpoints and remove duplicate segments to simplify SVG paths.
2) Convert vectors to 8K resolution binarized PNG while maintaining coordinate alignment.
3) Identify the largest connected component as the floor through connected component analysis, with secondary components as candidate walls.
4) Re-vectorize boundaries through coordinate mapping to generate closed wall polygons. 

The proposed method exhibits notable robustness against recognition inaccuracies – incomplete walls are automatically reconciled during stage 1, while small noise particles are filtered in stage 3 using area thresholds.

For doors and windows, we employ category-specific parameterization strategies.
Doors are systematically categorized into four distinct subtypes (single/double/sliding/folding) based on semantic labels, with positions, orientations, and pivot points determined through linear-arc spatial relationships.

Windows are grouped via union-find algorithms, calculating average lines of co-grouped segments as representative lines, then selecting the segment closest to the representative line within each group as the window instance position.

The final reconstruction leverages Blender's geometry nodes for procedural modeling: 
1) Extrude wall polygons to 3D volumes using floorplan height attributes. 2) Instance-place door/window assemblies at infered positions. 3) Apply floor based on origin CAD drawing.

This integrated approach achieves automated 3D reconstruction with exceptional efficiency, transforming raw CAD data into structurally complete 3D interior models within minutes through streamlined procedural generation.


\begin{table}[]
\centering
\setlength{\abovecaptionskip}{0pt}
\setlength{\belowcaptionskip}{0pt}
\begin{tabular}{c|ccc}
\hline
Primitive Pooling Type                      &PQ          & SQ          & RQ           \\ \hline
w/o Pooling  &81.9 & 89.4 & 91.6     \\ \hline
{Max}  & 86.9 & 93.3 & 93.2       \\ \hline
Average &  87.0 & 93.1 & \textbf{93.4}             \\ \hline
\textbf{Mixed (Max+Average)}   &\textbf{87.4}   & \textbf{93.5}   & \textbf{93.4}    \\ \hline
\end{tabular}
\caption{Ablation study on different pooling methods test on FloorPlanCAD dataset.}
\label{tab:pooling ablation}
\end{table}

\begin{table}[]
\centering
\begin{tabular}{c|ccc}
\hline
Method                       &PQ          & SQ          & RQ           \\ \hline
w/ color     &{87.4}   & {93.5}   & {93.3}        \\ \hline
w/o color      & 88.9 & 95.6 & 93.0        \\ \hline
\end{tabular}
\caption{Ablation study on with and without prior color imformantion as input. 
}
\label{tab:color ablation}
\end{table}

\begin{algorithm}
\caption{The pseudocode of SWA}
\label{alg:seg_fusion}
\begin{algorithmic}[1]
\Require sliding steps, CAD drawing, proposals $p$
\Ensure \texttt{final\_sem}, \texttt{final\_inst}
\State Initialize \texttt{tot\_sem} $\gets [N, C]$ \Comment{C: number of classes}
\State Initialize \texttt{tot\_inst} $\gets [\text{win\_num} \times p, N]$

\For{each sliding window $w_i$}
    \State \texttt{sem\_i} $\gets$ \Call{pred\_semantic}{$w_i$}
    \State \texttt{inst\_i} $\gets$ \Call{pred\_instance}{$w_i$}
    
    \For{each primitive Id $pId$}
        \State \texttt{sem\_i} $\gets$ \Call{one\_hot}{\texttt{sem\_i}[{$pId$}]}
        \State \texttt{tot\_sem}[{$pId$}] $\gets$ \texttt{tot\_sem}[{$pId$}] + \texttt{sem\_i}
    \EndFor
    \State \texttt{tot\_inst}[(i-1)*{$p$} : i*{$p$}] $\gets$ \texttt{inst\_i}
\EndFor
\State \texttt{final\_sem} $\gets$ \Call{argmax}{\texttt{tot\_sem}}
\State \texttt{final\_inst} $\gets$ \Call{NMS}{\texttt{tot\_inst}}
\end{algorithmic}
\end{algorithm}

\begin{table*}[htbp]
\centering
\begin{tabular}{c|ccc|ccc|ccc|ccc}
\hline
\multirow{2}{*}{\textbf{class}} & \multicolumn{3}{c|}{\textbf{A}}                                  & \multicolumn{3}{c|}{\textbf{B}}                        & \multicolumn{3}{c|}{\textbf{C}}                   & \multicolumn{3}{c}{\textbf{D}}                              \\ \cline{2-13} 
              & PQ    & SQ    & RQ    & PQ    & SQ    & RQ    & PQ    & SQ    & RQ    & PQ    & SQ    & RQ       \\ \hline
single door   & \textbf{91.90} & 96.33 & 95.40 & 84.11 & 91.34 & 92.08  & 87.23 & 92.68 & 94.12 & 87.46 & 92.85 & 94.20 \\
double door   & \textbf{94.10} & 96.46 & 97.55 & 87.23 & 91.70 & 95.13    & 87.30 & 91.85 & 95.04 & 88.27 & 92.73 & 95.18 \\
sliding door  & \textbf{96.50} & 98.02 & 98.46 & 92.88 & 95.04 & 97.73   & 94.51 & 96.15 & 98.29 & 91.32 & 95.65 & 95.46 \\
folding door  & \textbf{77.06} & 89.23 & 86.36 & 73.03 & 92.36 & 79.07  & {70.88} & 83.29 & 85.11 & 52.03 & 82.64 & 62.96 \\
revolving door  & 0.00 & 0.00 & 0.00 & 0.00 & 0.00 & 0.00 & 0.00 & 0.00 & 0.00 & 0.00 & 0.00 & 0.00 \\
rolling door  & 0.00 & 0.00 & 0.00 & 0.00 & 0.00 & 0.00 & 0.00 & 0.00 & 0.00 & 0.00 & 0.00 & 0.00 \\
window        & \textbf{85.41} & 91.06 & 93.80 & 77.71 & 84.43 & 92.04    & 77.13 & 85.24 & 90.49 & 74.53 & 85.06 & 87.62 \\
bay window     & \textbf{51.18} & 90.19 & 56.74 & 43.23 & 91.50 & 47.25 & 22.25 & 77.38 & 28.75 & 12.00 & 68.83 & 17.44 \\
blind window  & \textbf{86.00} & 92.25 & 93.22 & 77.99 & 86.97 & 89.68& 79.56 & 84.85 & 93.76 & 77.28 & 83.91 & 92.10 \\
opening symbol  & \textbf{52.67} & 77.06 & 68.35 & 34.51 & 74.91 & 46.07 & 36.47 & 79.11 & 46.10 & 20.10 & 72.22 & 27.84 \\
sofa          & \textbf{89.68} & 96.78 & 92.67 & 85.50 & 92.42 & 92.51   & 76.98 & 90.28 & 85.28 & 73.74 & 87.40 & 84.37 \\
bed           & \textbf{82.12} & 88.75 & 92.54 & 79.51 & 86.32 & 92.11   & 76.87 & 86.26 & 89.11 & 76.51 & 84.44 & 90.61 \\
chair         & \textbf{84.72} & 93.66 & 90.46 & 86.51 & 94.63 & 91.42   & {84.42} & 92.26 & 91.51 & 80.85 & 91.71 & 88.16 \\
table         & \textbf{77.42} & 89.58 & 86.42 & 73.30 & 87.39 & 83.87 & 68.03 & 86.46 & 78.68 & 61.99 & 86.43 & 71.72 \\
TV cabinet    & \textbf{94.56} & 97.08 & 97.40 & 83.23 & 85.80 & 97.01   & 89.99 & 93.41 & 96.33 & 79.94 & 85.50 & 93.49 \\
Wardrobe      & \textbf{91.56} & 94.54 & 96.85 & 88.97 & 92.63 & 96.05   & 85.37 & 88.77 & 96.17 & 82.50 & 85.94 & 95.99 \\
cabinet       & \textbf{80.10} & 89.04 & 89.95 & 73.11 & 84.25 & 86.78  & 71.37 & 82.66 & 86.33 & 69.15 & 83.42 & 82.90 \\
gas stove     & \textbf{97.89} & 99.21 & 98.67 & 96.76 & 98.39 & 98.35   & 97.10 & 97.89 & 99.19 & 96.66 & 97.61 & 99.03 \\
sink          & \textbf{87.08} & 93.71 & 92.93 & 84.87 & 91.49 & 92.77   & 84.88 & 90.98 & 93.30 & 83.42 & 90.03 & 92.65 \\
refrigerator  & \textbf{91.05} & 95.32 & 95.52 & 80.35 & 84.60 & 94.97   & 81.70 & 87.44 & 93.44 & 80.28 & 85.58 & 93.80 \\
air conditioner  & \textbf{87.43} & 98.17 & 89.06 & 82.21 & 95.67 & 85.93   & 80.59 & 92.04 & 87.56 & 75.45 & 91.70 & 82.27 \\
bath          & \textbf{75.50} & 88.30 & 85.50 & 69.96 & 81.78 & 85.54 & 66.62 & 79.71 & 83.58 & 62.06 & 77.36 & 80.22 \\
bath tub      & \textbf{83.67} & 88.14 & 94.92 & 76.50 & 81.37 & 94.01   & 69.94 & 80.14 & 87.27 & 64.82 & 78.13 & 82.96 \\
washing machine  & \textbf{88.26} & 95.38 & 92.54 &82.34 & 88.59 & 92.95   & 86.40 & 91.21 & 94.73 & 77.97 & 85.26 & 91.45 \\
squat toilet  & \textbf{92.85} & 96.01 & 96.71 & 92.41 & 96.23 & 96.03   & 92.58 & 94.79 & 97.67 & 89.99 & 92.68 & 97.09 \\
urinal        & \textbf{94.46} & 97.76 & 96.62 & 91.24 & 94.17 & 96.90   & 91.80 & 94.84 & 96.79 & 91.31 & 94.79 & 96.32 \\
toilet        & \textbf{95.83} & 97.70 & 98.09 & 91.22 & 92.96 & 98.13   & 90.81 & 93.01 & 97.63 & 90.48 & 93.53 & 96.74 \\
stairs        & \textbf{84.41} & 92.37 & 91.38 & 80.19 & 88.51 & 90.60   & 68.56 & 82.02 & 83.59 & 65.50 & 79.52 & 82.36 \\
elevator      & \textbf{95.06} & 97.57 & 97.43 & 93.26 & 96.46 & 96.69   & 84.44 & 90.52 & 93.28 & 80.64 & 88.59 & 91.03 \\
escalator     & \textbf{74.96} & 87.62 & 85.55 & 70.43 & 85.84 & 82.05   & 29.80 & 72.72 & 40.98 & 44.64 & 73.70 & 60.56 \\
row chairs    & \textbf{85.73} & 94.58 & 90.65 & 84.15 & 93.96 & 89.55  & {85.69} & 94.53 & 90.65 & 84.45 & 93.99 & 89.86 \\
parking spot  & \textbf{88.54} & 93.15 & 95.05 & 84.48 & 90.24 & 93.62   & 72.22 & 80.84 & 89.34 & 72.64 & 82.77 & 87.76 \\
wall        & \textbf{73.21} & 81.42 & 89.91 & 62.55 & 73.32 & 85.31   & 44.88 & 65.29 & 68.74 & 44.48 & 65.44 & 67.97 \\
curtain wall  & 58.66 & 82.61 & 71.01 & 50.49 & 78.00 & 64.73   & \textbf{60.25} & 87.36 & 73.90 & 37.62 & 73.21 & 51.39 \\
railing       & \textbf{67.25} & 88.38 & 76.10 & 59.50 & 83.88 & 70.94   & 37.82 & 76.42 & 49.49 & 28.53 & 70.73 & 40.34 \\\hline
total         &\textbf{87.36}   & {93.40}   & {93.54}   & 81.89 & 89.42 & 91.58 & 80.24 & 90.17 & 88.98 & 78.36 & 88.59 & 88.46 \\                     
\hline
\end{tabular}
\caption{Quantitative results of the ablation study for panoptic symbol spotting across different classes.\\
   A: Dense point sampling + PTv3 + Pooling(Ours). B: Dense point sampling + PTv3. C: Handcrafted features + PTv3. D: Handcrafted features + PTv1}
\label{tab:class_PQ}
\end{table*}

\begin{figure*}
  \centering
  \setlength{\abovecaptionskip}{0pt}
  \setlength{\belowcaptionskip}{0pt}=
   \includegraphics[width=1\linewidth]{{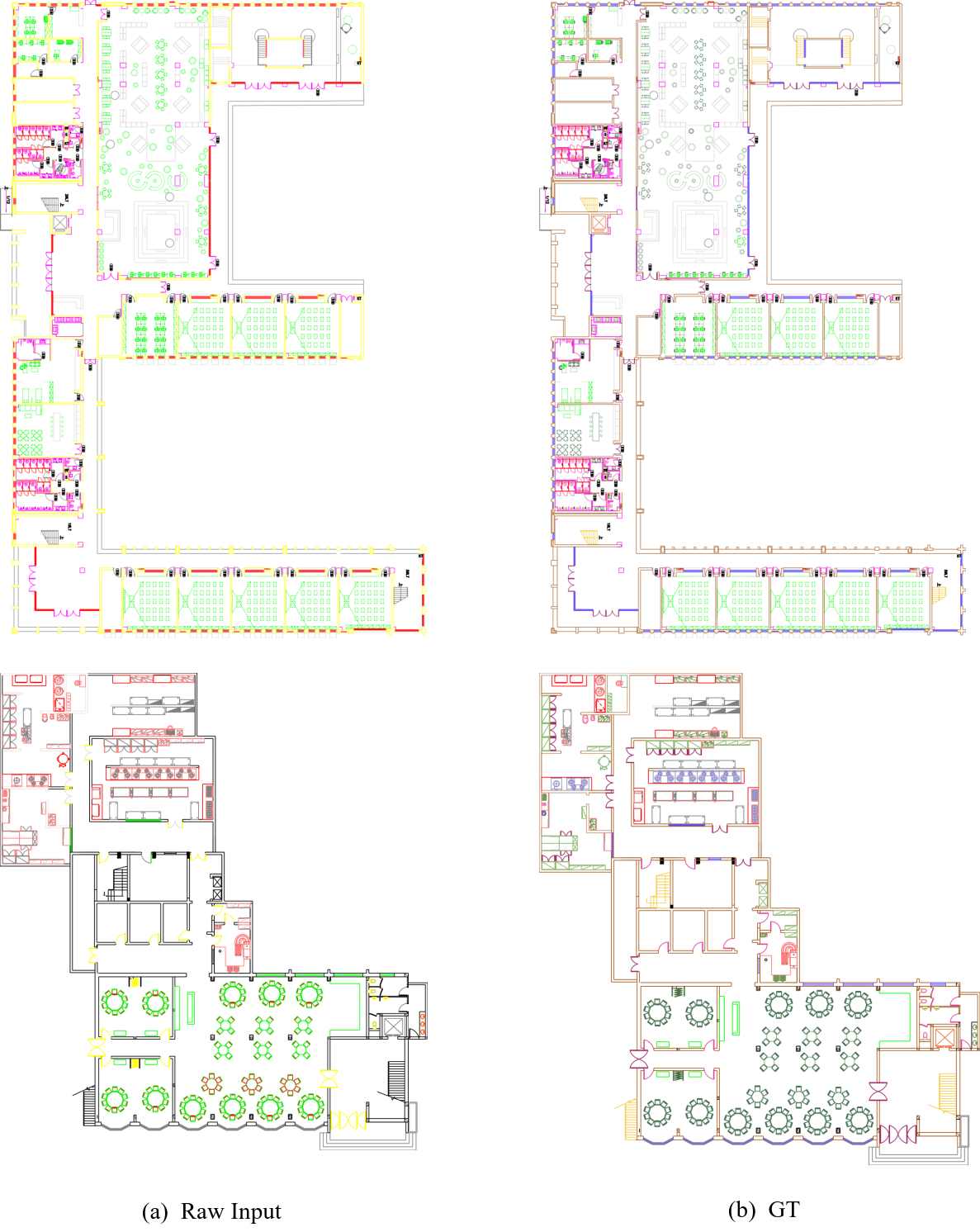}}
   \caption{Examples from the LS-CAD Dataset. The figure shows the raw input and ground truth of two samples in the dataset. The raw input represents the unprocessed original svg image, while the ground truth is the svg image generated based on manually annotated labels, which facilitates the visualization and comparison of results. This dataset is designed to assess the performance of CAD image processing algorithms.}
   \label{fig:LS_raw_gt}
\end{figure*}

\begin{figure*}
  \centering
  \setlength{\abovecaptionskip}{0pt}
  \setlength{\belowcaptionskip}{0pt}
   \includegraphics[width=1\linewidth]{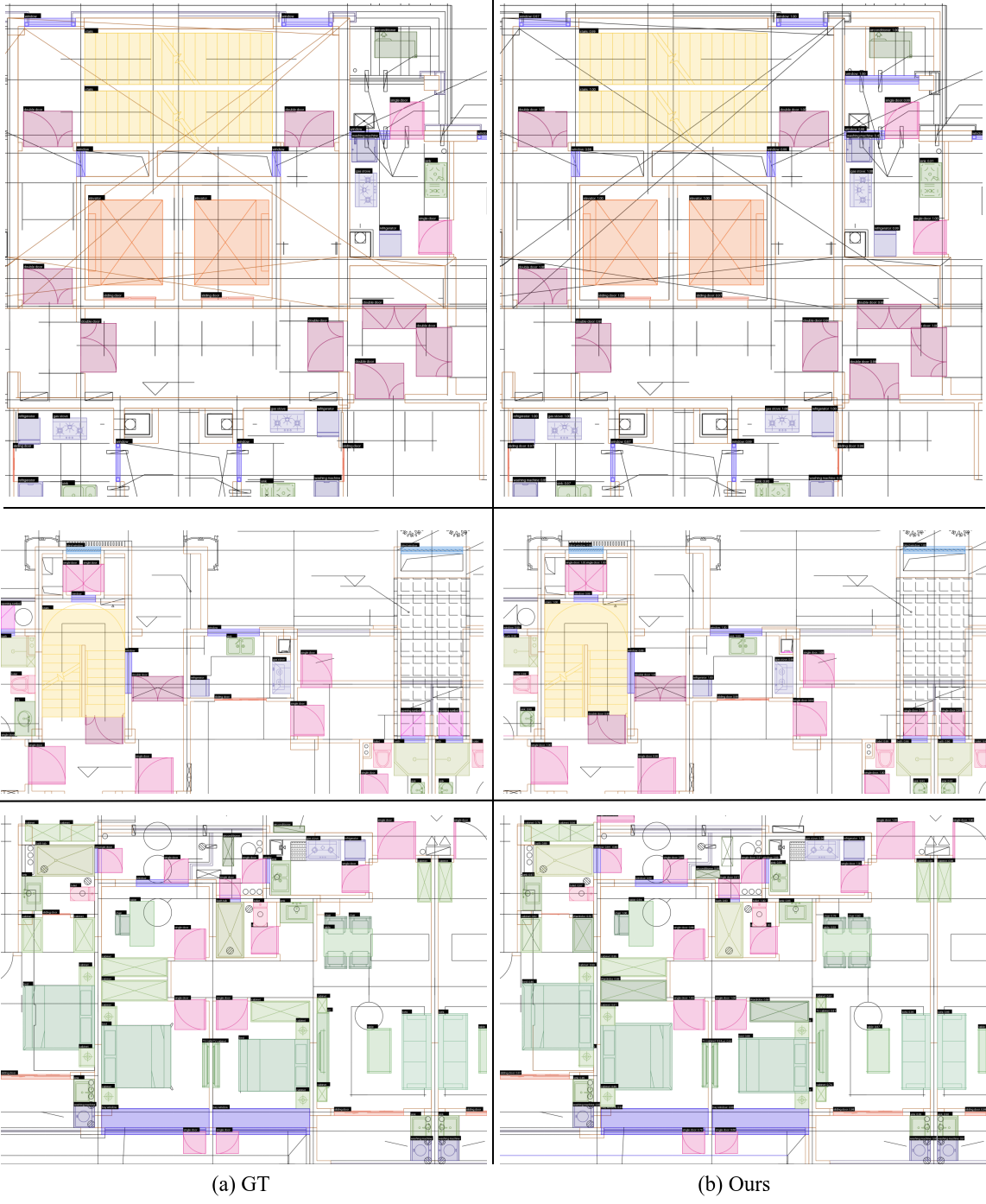}

   \caption{Qualitative comparison of panoptic symbol spotting. 
   Our method accurately detects symbol instances, even in situations where symbols are overlapped by other elements.}
   \label{fig:qualitative_comparison_1}
\end{figure*}

\begin{figure*}
  \centering
  \setlength{\abovecaptionskip}{0pt}
  \setlength{\belowcaptionskip}{0pt}
   \includegraphics[width=1\linewidth]{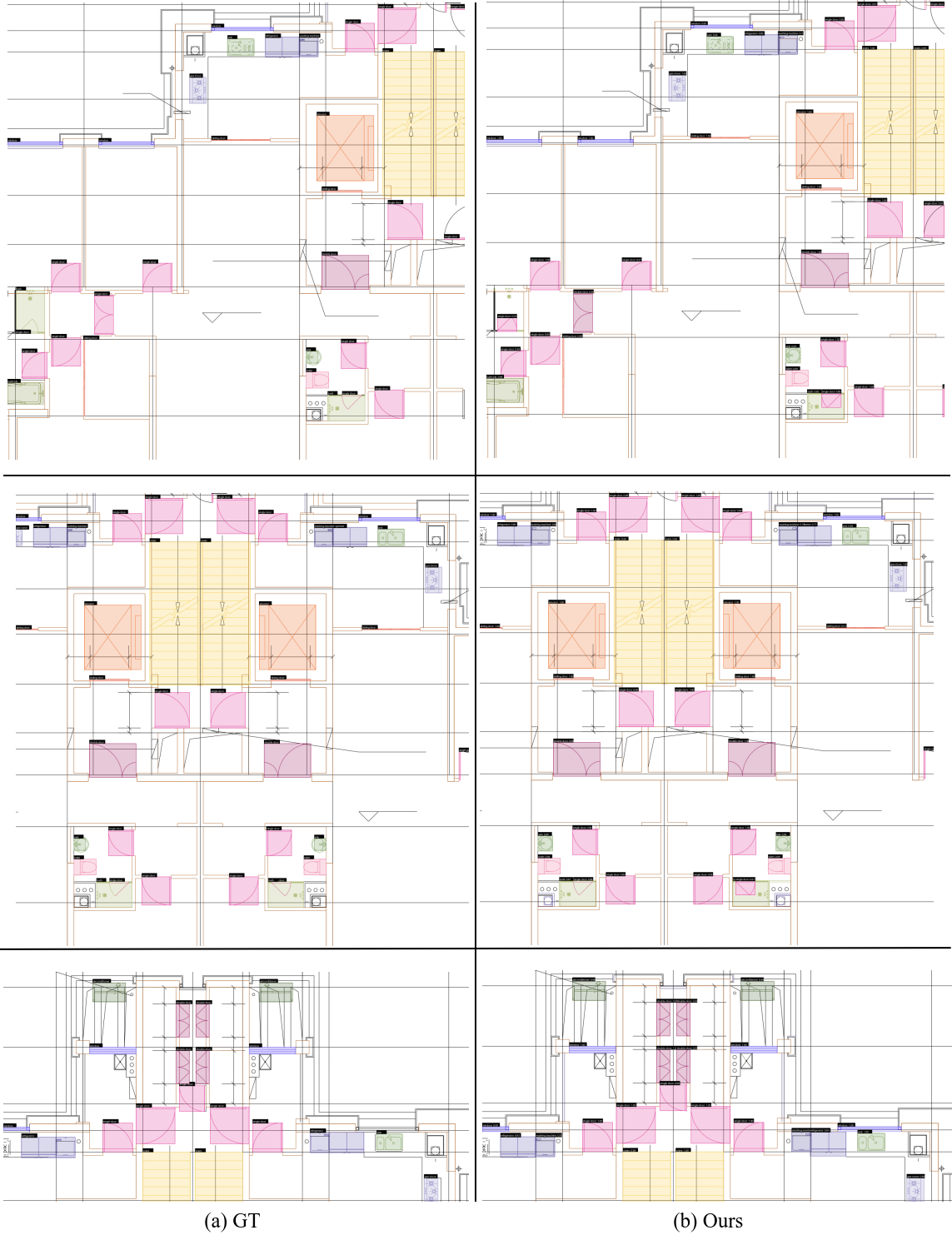}

   \caption{Qualitative comparison of panoptic symbol spotting. 
   Our method accurately detects symbol instances, even in situations where symbols are overlapped by other elements.}
   \label{fig:qualitative_comparison_2}
\end{figure*}

\begin{figure*}
  \centering
  \setlength{\abovecaptionskip}{0pt}
  \setlength{\belowcaptionskip}{0pt}
   \includegraphics[width=1\linewidth]{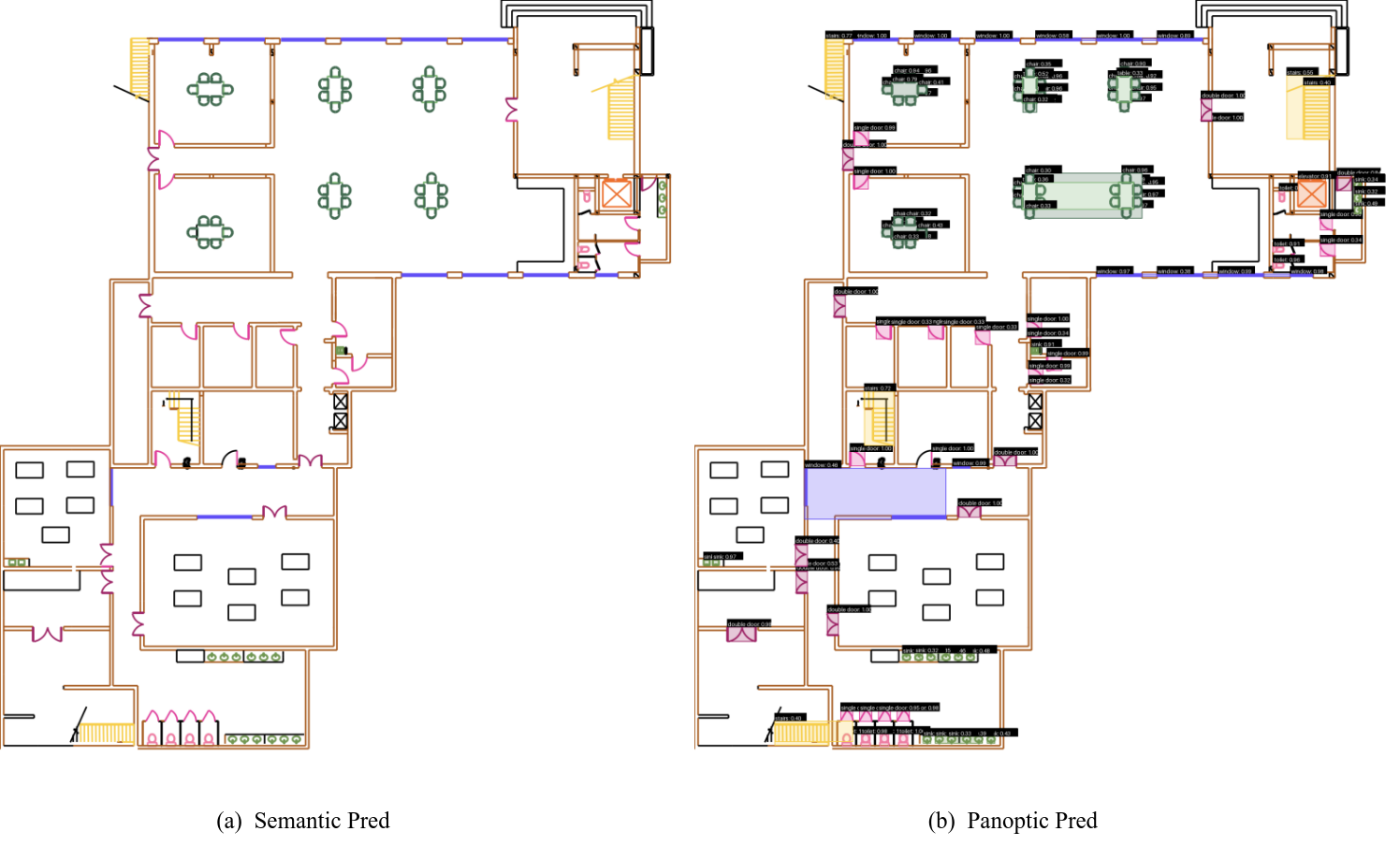}

   \caption{Qualitative visualization highlighting the performance of our method for panoptic symbol spotting on the LS-CAD dataset.}
   \label{fig:qualitative_comparison_3}
\end{figure*}

\begin{figure*}
  \centering
  \setlength{\abovecaptionskip}{0pt}
  \setlength{\belowcaptionskip}{0pt}
   \includegraphics[width=1\linewidth]{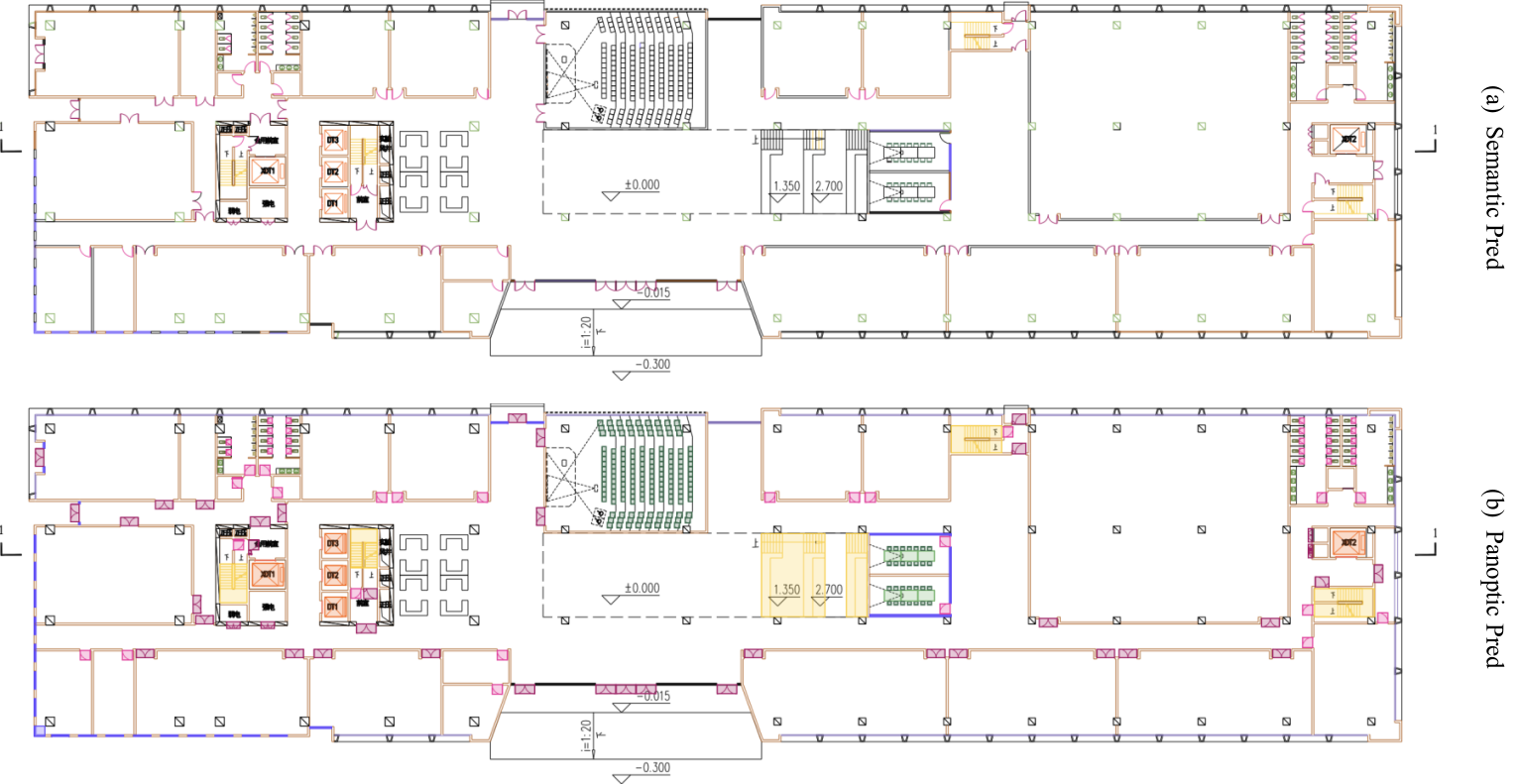}

   \caption{Qualitative visualization highlighting the performance of our method for panoptic symbol spotting on the LS-CAD dataset.}
   \label{fig:qualitative_comparison_4}
\end{figure*}

\begin{figure}[t]
  \centering
  \setlength{\abovecaptionskip}{0pt}
\setlength{\belowcaptionskip}{0pt}
   \includegraphics[width=1\linewidth]{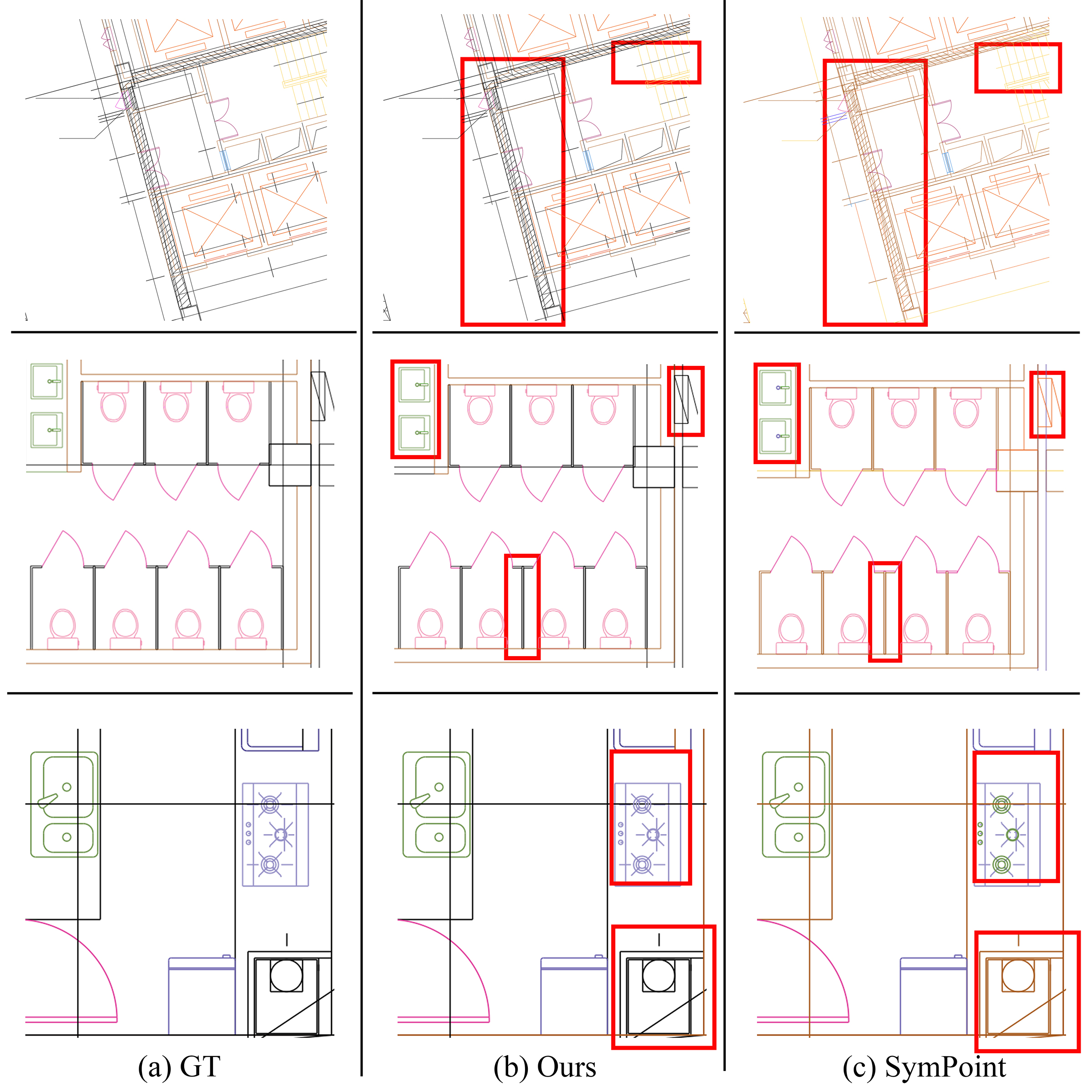}

   \caption{Qualitative comparison of semantic symbol spotting. Our dense point sampling based primitive feature learning enables our approach to achieve precise semantic segmentation, particularly on walls and complex, overlapping symbols such as faucets in sinks and uncommon furniture symbols.}
   \label{fig:sem_comparison}
\end{figure}

\begin{figure}[t]
  \centering
  \setlength{\abovecaptionskip}{0pt}
  \setlength{\belowcaptionskip}{0pt}
  \begin{subfigure}[b]{1\linewidth}
    \includegraphics[width=1\linewidth]{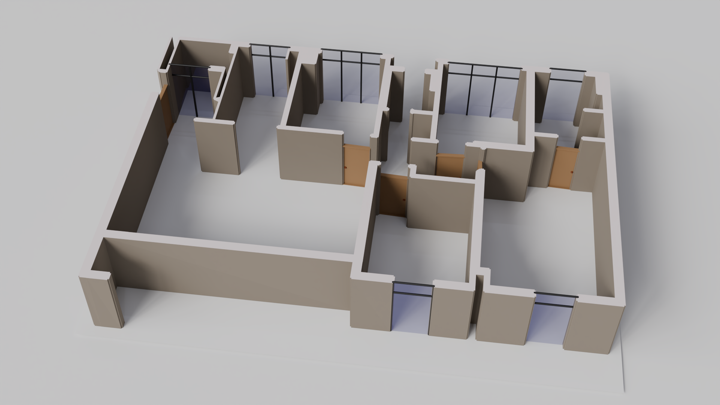}
   \caption{Automated 3D reconstruction of small-scale architectural building.}
   \label{fig:house003}
  \end{subfigure}
  
  \vspace{2mm} 
  
    \begin{subfigure}[b]{1\linewidth}
    \includegraphics[width=1\linewidth]{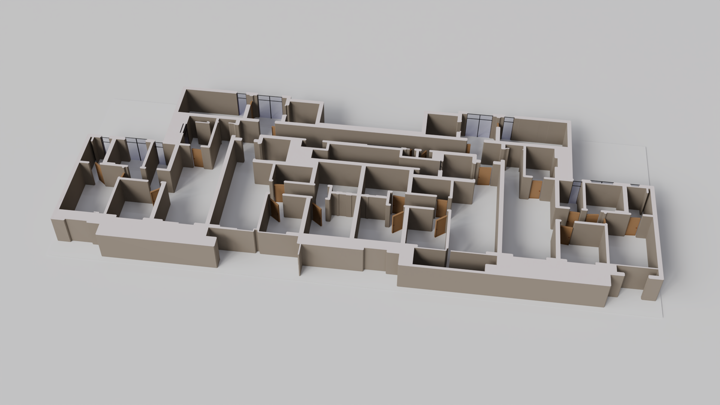}
   \caption{Automated 3D reconstruction of large-scale architectural building.}
   \label{fig:house001}
    \end{subfigure}
    
  \caption{Automated 3D reconstruction results.}
  \label{fig:modeling results}
\end{figure}
\end{document}